%% file: main.tex
\documentclass[11pt]{article}
\usepackage{elorian}
\usepackage[round,authoryear]{natbib}

\usepackage[T1]{fontenc}
\usepackage[utf8]{inputenc}
\usepackage{microtype}

\usepackage{amsmath,amssymb,amsthm}
\usepackage{graphicx}
\graphicspath{{figures/}}
\usepackage{booktabs}
\usepackage{multirow}
\usepackage{siunitx}
\usepackage{xcolor}
\usepackage{pifont}
\usepackage{mdframed}
\usepackage{placeins}
\usepackage{float}

\usepackage{url}
\usepackage{hyperref}
\usepackage{cleveref}
\hypersetup{
  colorlinks=true,
  linkcolor=blue!55!black,
  citecolor=blue!55!black,
  urlcolor=blue!55!black,
  pdftitle={DecoyCount: Evaluating Visual-Reasoning Misalignment in Counting},
  pdfauthor={Qiuyi (Richard) Zhang, Zhen Xu, Jing Gu, Seojin Bang, Arnav Mohanty Das, Yaning Hu, Gautam Kumar, Jared Lichtarge, Le Xue, Bettina Catoto, Akul Datta, Daniel Luo, Farha Khan, Marcella Valentine, Yinfei Yang, and Andrew M. Dai}
}

\newcommand{\cmark}{\ding{51}}

\newcommand{\safegraphic}[2][]{%
  \IfFileExists{#2}{\includegraphics[#1]{#2}}{%
    \fbox{\parbox[c][2.4cm][c]{0.82\linewidth}{\centering\ttfamily\scriptsize
      MISSING FIGURE ASSET\\[2pt]\detokenize{#2}}}}}
\title{DecoyCount: Evaluating Visual-Reasoning Misalignment in Counting}
\author{Qiuyi (Richard) Zhang$^{1}$, Zhen Xu$^{1}$, Jing Gu$^{1}$, Seojin Bang$^{1}$,
Arnav M. Das$^{1}$, Yaning Hu$^{1}$, Gautam Kumar$^{1}$, Jared Lichtarge$^{1}$,
Le Xue$^{1}$, Bettina Catoto$^{2}$, Akul Datta$^{2}$, Daniel Luo$^{2}$, Farha Khan$^{2}$,
Marcella Valentine$^{1}$, Yinfei Yang$^{1}$, Andrew M. Dai$^{1}$}
\renewcommand{\paperaffiliations}{$^{1}$Elorian AI \qquad $^{2}$Mercor}

\newcommand{\dataset}{\textnormal{\textsc{DecoyCount}}}
\newcommand{\datasetmh}{\textnormal{\textsc{DecoyCount}} multi-hop}
\newcommand{\rgap}{\Delta R}
\newcommand{\rgapp}{\Delta^{2}R}
\definecolor{gapgreen}{RGB}{0,105,75}
\definecolor{gapred}{RGB}{175,45,45}
\newcommand{\gapnarrows}[1]{\textcolor{gapgreen}{\ensuremath{\mathbf{#1}}}}
\newcommand{\gapwidens}[1]{\textcolor{gapred}{\ensuremath{\mathbf{#1}}}}

\begin{document}
\maketitle
\pagestyle{plain}

\begin{abstract}

Visual counting benchmarks report whether vision-language models produce the correct number, but
not whether errors arise from perception or reasoning. We introduce \mbox{\dataset{}}, a benchmark
built from 1,586 real-world images with designed decoys across seven mechanisms, along with paired prompts: a neutral question, a matched alignment cue, and a novel multi-hop, multi-image question that situates the same counting problem in a realistic scenario. Using this design, we find that counting failures are induced partly by visual misalignment: models can execute the intended counting process
when cue-guided but fail to select it unaided, and we term this difference in accuracy the \emph{reasoning-alignment gap}. Despite being highly capable on general reasoning, frontier models remain below \SI{50}{\percent} accuracy on our benchmark, while alignment gaps range from
12.1--22.1 percentage points; strikingly, we observe that more capable models, greater thinking budgets, and agentic processes
often induce larger gaps. Because humans may bring context that models lack, we test whether textual
underspecification contributes to misalignment using a scenario-based multi-hop condition without explicit hinting, but this added context does not improve aggregate performance. Our analysis
concludes that stronger general reasoning does not by itself generalize to aligned visual reasoning,
leaving misalignment with human visual expectations as a substantial component of the remaining deficit.
\dataset{} and the evaluation code are released at
\url{https://github.com/elorian-ai/decoycount}.
\end{abstract}

\input{sec1_introduction}
\FloatBarrier
\input{sec2_related_work}
\input{sec3_dataset}
\input{sec4_setup_and_models}
\input{sec5_main_findings}
\FloatBarrier
\bibliographystyle{plainnat}
\bibliography{references}

\newpage
\appendix
\input{appendix}

\end{document}

%% file: sec1_introduction.tex
\section{Introduction}
\label{sec:intro}
Visual counting appears elementary, but a correct answer requires a model to locate the target,
decide what constitutes one instance and the requested unit, distinguish physical objects from
reflections, depictions, or printed numerals, and enumerate the resulting set. Each step can fail for
a different reason, yet a single final number collapses them into one score. Standard counting
accuracy therefore cannot distinguish a model that fails to perceive an object from one that sees
the scene but selects an unintended interpretation or reasoning shortcut.

Most reasoning benchmarks emphasize explicit chains of logical or mathematical inference: if $A$
implies $B$, can a model carry that implication through multiple steps? Reasoning is broader,
however; it also includes causal and analogical inference, forming and revising hypotheses, creative
problem solving, planning, and selecting the strategy appropriate to a situation. Existing
benchmarks rarely separate these facets, especially when reasoning is grounded in images. It is
therefore difficult to tell whether a failure actually reflects limited reasoning capability.

\paragraph{Our Approach.}

\begin{figure}[t]
\centering
\safegraphic[width=\linewidth]{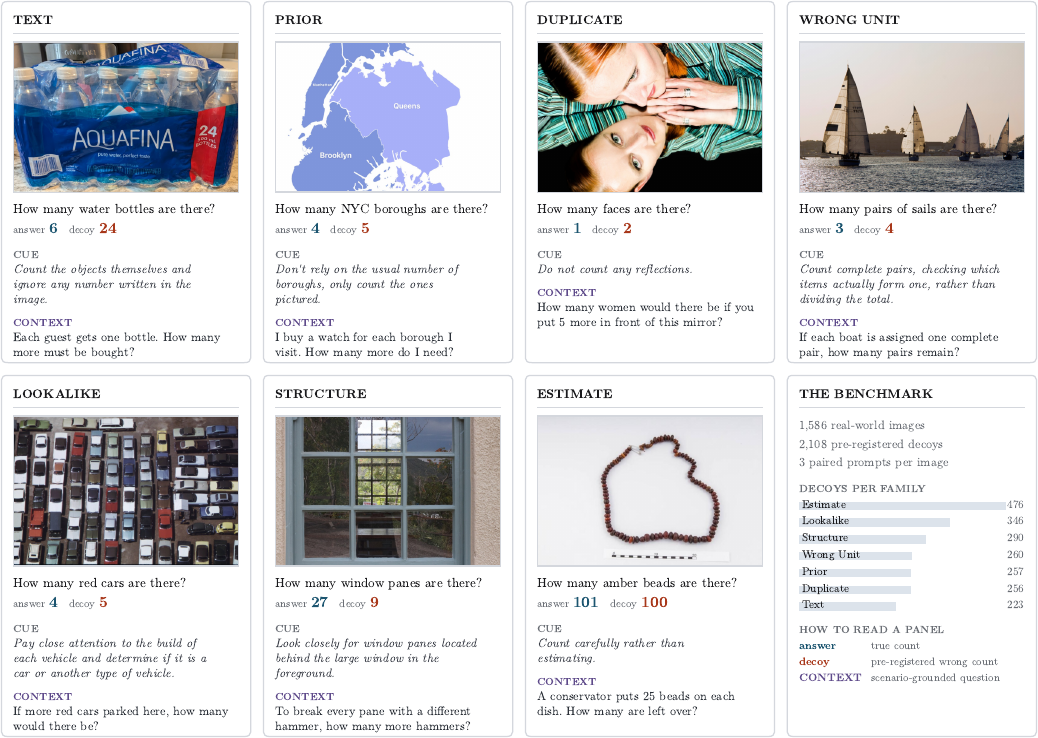}
\caption{\textbf{DecoyCount exposes interpretable counting failures across seven mechanisms.}
One item per family: the neutral question with its ground-truth count (blue) and pre-registered
decoy (red), the alignment cue (italic) naming the intended counting process without revealing the
answer, and an abridged contextual question (purple) that implies the rule through a scenario
instead. Most contextual items also supply a second image, not shown here. The final panel
summarizes the 2,108 decoys over 1,586 images.}
\label{fig:examples}
\end{figure}

We introduce \dataset{}, a diagnostic benchmark with real-world counting items and pre-registered
decoys from seven mechanisms (\Cref{fig:examples}). A \emph{decoy} records the plausible wrong
count suggested by one mechanism, revealing which alternative interpretation a model follows. For
example, when six bottles sit in packaging printed with ``24,'' the printed number is the decoy; in
a mirrored scene, the reflection can suggest a doubled count. Each item pairs a neutral question
with a matched \emph{alignment cue} that specifies the intended process without revealing the
answer, such as ``Ignore numbers written in the image'' or ``Do not count mirrored objects.'' In addition to a decoy taxonomy, we source these images via a scene-based taxonomy to ensure diversity and remove images that are excessively ambiguous.

From these paired prompts, we introduce and quantify the \textbf{reasoning-alignment gap}
($\rgap$), defined as cued accuracy minus neutral accuracy. A positive gap measures the average
accuracy gained when the intended counting process is specified while the image and target
answer remain fixed. Calling this an alignment gap may seem surprising because counting appears to
have a uniquely determined visual answer. However, two observations motivate the term: minimal cues shift
accuracy by 12.1--22.1 percentage points, and changing only ``are there'' to ``are visible''
shifts Gemini-3.8-flash accuracy by 6.69 points. The intended process therefore depends partly on human
interpretation. The reasoning-alignment gap isolates one component of visual-reasoning misalignment
rather than all perceptual or reasoning limitations.

Because real-world images may invite multiple interpretations, we define alignment
relative to the interpretation implied by the most common real-world scenarios in which the objects
appear or are used. For example, ``How many apples are there?'' may leave a reflected
apple ambiguous, whereas a pie-making scenario establishes that only the physical apple is usable.
To test whether such underspecification causes the gap, we add a paired multi-hop version of each
question that grounds the same count in a realistic scenario without naming the counting rule, with
a second image when required to match the complexity of real-world counting queries. To our
knowledge, this is the first counting benchmark to pair the same underlying count with
scenario-based, multi-hop, multi-image questions. Interestingly, adding this grounding scenario does not improve
aggregate performance and rescues only \SIrange{11.9}{32.7}{\percent} of neutral failures per
family; underspecification therefore contributes to some errors but does not explain the full gap.

Across six frontier models, neutral accuracy reaches to only 43.2\%, while adding alignment cues
produce reasoning gaps of 12.1--22.1 percentage points and remove \SIrange{43.4}{65.8}{\percent} of exact decoy hits. We test several alternative
explanations for this effect. If stronger capability aligned the default reasoning process, scale
and thinking should shrink the gap; instead, Qwen3.5 4B$\to$27B raises neutral accuracy by 5.30
points while widening the gap by 4.52, and native Qwen3.8 thinking raises accuracy from
\SI{12.65}{\percent} to \SI{27.17}{\percent} while the gap grows from 8.96 to 20.80 points. If the
cue merely elicited more effort, generic or mismatched cues should reproduce the gain; they reproduce at
most \SI{20}{\percent}. If benchmark construction caused the effect, editing should change the gap
and standard model rankings should break; instead, editing primarily changes difficulty and the
rankings persist. The mechanism split further supports the diagnosis: Estimate is predominantly
consistent with a perceptual or enumeration bottleneck, while five families exhibit large reasoning-alignment gaps. Finally, if tools supply missing
visual evidence, targeted instrumentation should help; general CI agentic harnesses widen the gap by
2.48--7.17 points, whereas SAM3 narrows it by 4.52. Together, these results identify
visual-reasoning misalignment as a substantial but incomplete source of error: strong general
reasoning does not ensure that a model selects the intended visual reasoning process. Residual
errors may still reflect perception or other reasoning limitations.

\paragraph{Our Contributions.} Here are our main contributions, summarized:
\begin{enumerate}[leftmargin=*,itemsep=1pt,topsep=2pt]
  \item \textbf{A paired, scenario-grounded benchmark with interpretable decoys.} \dataset{} pairs
  1,586 real-world items with neutral questions, mechanism-specific decoys, and answer-free
  alignment cues. Each decoy records a plausible wrong count induced by one of seven mechanisms;
  its matched cue identifies the intended reasoning process without revealing the answer. Paired
  multi-hop questions ground that intended count in common real-world scenarios, including 783
  multi-image items.
  \item \textbf{We identify and quantify a wide reasoning-alignment gap.} We define the
  \emph{reasoning-alignment gap} as cued accuracy minus neutral accuracy on the same image and target
  answer, measuring the accuracy gained when the intended counting process is specified.
  Across frontier models, neutral accuracy is
  \SIrange{17.1}{43.1}{\percent}, and five of six models return the decoy more often than the truth.
  Matched cues add 12.1--22.1 percentage points, repair 2.7--6.8$\times$ more answers than they
  break, and remove \SIrange{43.4}{65.8}{\percent} of decoy hits. The large reasoning-alignment gap highlights a novel and significant component of reasoning failures.
  \item \textbf{Additional context does not close the gap.} Although the multi-hop
  scenarios ground the intended count without naming the counting rule, they do not improve
  aggregate performance and rescue only \SIrange{11.9}{32.7}{\percent} of failures per family.
  Missing or minimized context therefore explains some individual failures, but not the full misalignment.
  \item \textbf{Greater capability does not ensure alignment.} Across
  controlled scale and thinking efforts sweeps, we find that accuracy improves while the gap widens by 4.52 and 11.84
  points, respectively. Therefore, more capable models are better at general reasoning but this gain does not translate into correct visual alignment. We find that general CI (Code Interpreter) agentic harnesses also follow a similar trend, improving accuracy but widening the gap by 2.48--7.17 points.
\end{enumerate}

%% file: sec2_related_work.tex
\section{Background and Related Work}
\label{sec:related}
\Cref{tab:related} compares counting, adversarial, counterfactual, and multi-hop benchmarks across
five diagnostic design axes. Existing benchmarks typically specialize in only a subset of them.

\begin{table}[t]
\centering
\scriptsize
\setlength{\tabcolsep}{2.2pt}
\begin{tabular}{@{}lccccc@{}}
\toprule
\textbf{Benchmark} & \shortstack{\textbf{Image Decoy}} & \textbf{Taxonomy} & \textbf{Cue} & \textbf{Multi-hop} & \textbf{Multi-image} \\
\midrule
TallyQA \citeyearpar{acharya2019tallyqa}             & ---                   & simple / complex only     & ---     & ---                    & --- \\
CountBench \citeyearpar{paiss2023countbench}         & ---                   & ---                       & ---     & ---                    & --- \\
PixMo-Count \citeyearpar{deitke2025molmo}            & ---                   & ---                       & ---     & ---                    & --- \\
CountQA \citeyearpar{tamarapalli2025countqa}         & ---                   & ---                       & ---     & ---                    & --- \\
VLMCountBench \citeyearpar{guo2025vlmcount}          & ---                   & by composition level      & generic & ---                    & --- \\
HoloCount \citeyearpar{deng2026holocount}            & ---                   & 20 skill subsets, 3 tiers & ---     & comparison only        & --- \\
\midrule
VLMBias \citeyearpar{vo2025biased}                   & \cmark\ (prior only)  & by domain                 & generic & ---                    & --- \\
CounterCount \citeyearpar{alzahrani2026countercount} & \cmark\ (prior only)  & 10 semantic categories    & ---     & ---                    & --- \\
\midrule
HotpotQA / MuSiQue                                   & ---                   & bridge vs.\ comparison    & ---     & \cmark                 & --- \\
WebQA / MMIU / MultihopSpatial                       & ---                   & ---                       & ---     & \cmark\ (not counting) & \cmark \\
\midrule
\dataset{} \textbf{(ours)} & \textbf{\cmark\ 2,108} & \textbf{7 mechanisms} & \textbf{\cmark\ per item} & \textbf{\cmark\ paired} & \textbf{\cmark\ 783 items} \\
\bottomrule
\end{tabular}
\caption{\textbf{DecoyCount combines adversarial multimodal decoys, alignment cues, and realistic multi-hop, multi-image
counting.} Prior benchmarks cover individual elements; \dataset{} pairs 2,108 pre-registered decoys
across seven mechanisms with per-item alignment cues and bridge questions, including 783 multi-image
items. ``Prior only'' denotes a canonical-count decoy.}
\label{tab:related}
\end{table}

\paragraph{Counting benchmarks.} The largest counting sets ask for totals over crowds, cells or
repeated retail items with dense supervision inherited from detection
\citep{zhang2016crowd,ranjan2021counting}; VLMCountBench \citep{guo2025vlmcount} pushes the same axis
to synthetic geometry, where models count one shape type reliably and fail once several are mixed.
Language-posed variants do the same in QA form: TallyQA \citep{acharya2019tallyqa} at roughly
\num{287000} questions, CountBench \citep{paiss2023countbench} on image--text count agreement,
PixMo-Count \citep{deitke2025molmo} with pointing supervision, and CountQA
\citep{tamarapalli2025countqa} on cluttered real photographs, where the best of fifteen MLLMs reaches
\SI{42.9}{\percent} when the benchmark was released. HoloCount \citep{deng2026holocount} is the most diagnostically structured: 2480
questions over 20 subsets in a three-level taxonomy whose robustness tier includes counter-prior
items, visual distractors, null targets and dense fields. These benchmarks primarily report
aggregate or subset accuracy rather than item-specific failure mechanisms. 

\paragraph{Adversarial and hard benchmarks.} Benchmarks must remain challenging as models improve,
and many create diagnostic failures
either by selecting examples that fool a model or by making controlled edits that change the correct
label. HellaSwag uses adversarial filtering, in which discriminators iteratively select difficult
machine-generated wrong endings \citep{zellers2019hellaswag}; contrast sets make small,
meaning-changing edits to expose local decision boundaries \citep{gardner2020contrast}. In visual
counting, VLMBias \citep{vo2025biased} edits familiar subjects across seven domains so the visible
count conflicts with the canonical count, yielding \SI{17.05}{\percent} average accuracy and a
\SI{75.7}{\percent} prior-consistent bias rate. CounterCount \citep{alzahrani2026countercount} pairs
168 factual-condition images with 168 counterfactual edits, localizes the relevant evidence, and
finds that models may attend to it yet underuse it. Both counting benchmarks isolate canonical-count
priors.


\paragraph{Multi-hop reasoning.} Complex, realistic reasoning often requires multiple steps, commonly
organized in two structures: a \emph{bridge} question feeds one hop's output into the next
\citep{yang2018hotpotqa}, whereas a \emph{comparison} question combines independent lookups;
MuSiQue reduces shortcuts that bypass the intended chain \citep{trivedi2022musique}. Multimodal work
inherits this distinction: WebQA chains retrieval across text and images, ReasonVQA combines
grounding with knowledge, MultihopSpatial builds 4500 spatial queries, and MMIU spans 52 multi-image
tasks \citep{chang2022webqa,tran2025reasonvqa,lee2026multihopspatial,meng2025mmiu}; none targets
counting. Existing counting composition is chiefly comparative: HoloCount's ``how many more $X$ than
$Y$?'' subtracts two independent counts.


%% file: sec3_dataset.tex
\section{The \dataset{} Benchmark}
\label{sec:decoycount}
\dataset{} is a paired benchmark built around registered decoys from seven counting-error
mechanisms. Each neutral question has a corresponding alignment cue that targets its decoy mechanism
and a scenario-grounded multi-hop question whose answer depends on the same visual count. Because all
three conditions depend on the same underlying count, their differences separate the effects of
targeted guidance and contextual grounding, exposing underlying model reasoning misalignment.

\subsection{Tasks and Decoys}
\label{sec:problem}
\label{sec:stats}
\label{sec:tasks}
Each item contains an image, a counting question, the true count, and one or more pre-registered
\emph{decoys}: wrong counts invited by the scene. The neutral and cued conditions use the same image
and question; their difference defines the reasoning-alignment gap while holding the underlying
visual difficulty. The paired \datasetmh{} condition places the same source image, and for realism sometimes an additional second image, inside
a real-world multi-hop bridge question. Although naturally more complex, it serves as a contextual
grounding control rather than simply a harder reasoning tier, testing whether missing real-world
context contributes to ambiguity in the neutral question without supplying a matched alignment cue.
For each contextual item, the final answer is not shared with the neutral question due to its multihop nature but applying the bridge question to the decoy yields an incorrect answer.

Seven families describe how a scene suggests a wrong count: \textbf{Lookalike}, \textbf{Estimate},
\textbf{Prior}, \textbf{Structure}, \textbf{Duplicate}, \textbf{Text}, and \textbf{Wrong Unit}.
The taxonomy was fixed before collection and balanced for per-family analysis. A second, independent axis records the kind of scene: seven mutually exclusive domains covering
everyday, public and outdoor, document and artwork, chart and scientific, retail, observation, and
industrial imagery. Sampling holds each domain below \SI{21}{\percent} of the release and places
every family in all seven, so a per-family score is not a proxy for one recurring setting.
Per-family and per-domain composition is in \Cref{tab:families}.

\paragraph{Decoy and alignment-cue design.}
The families intervene at different stages of counting. Lookalike, Duplicate, and Wrong Unit alter
which instances belong in the counted set; Text and Prior provide competing answers; Structure and
Estimate challenge enumeration itself. Annotators pre-register the target concept, true count,
integer decoys, and their visible mechanisms. Each decoy must be supported by the image and differ
from the truth; only items eliciting at least one such hit from a gating model are retained.
Pre-registration makes the hit evidence of a specified shortcut rather than an arbitrary error, and
all qualifying labels are kept when mechanisms overlap.

A matched alignment cue minimally states the relevant counting rule without changing the image,
target category, or answer space, and never reveals the correct count. Depending on the family, it
may direct the model to count physical instances, ignore printed numerals, use the requested unit,
verify a repeated structure, or enumerate rather than estimate. Because the cue targets the
registered mechanism rather than adding visual evidence, the reasoning-alignment gap is informative
about process selection. Generic and mismatched controls in \Cref{sec:robustness} show that the gap
does not follow from any added instruction or reasoning effort. Together, pre-registered decoys and matched
cues connect the taxonomy in \Cref{fig:examples,tab:taxonomy} to the reasoning-alignment gap.

\subsection{Collection and Validation}
\label{sec:collection}
\label{sec:annotation}
Images combine commissioned photography with public-domain and openly licensed material; some are
edited or generated when the intended mechanism cannot be found naturally. Provenance and licence
fields are recorded at authoring. Each item records its target, true count, up to five decoys and
their visible mechanisms, three paired prompts, and an optional second image. Candidates then pass a three-model gate using Claude Opus 5, GPT-5.6 Sol, and Gemini 3.6 Flash. The
difficulty gate requires all three to miss the neutral answer; the decoy-quality gate requires at
least one to return a registered decoy exactly.

Independent reviewers then verify that the target is countable, visually grounded, and unambiguous.
Neutral and contextual tasks are counted blindly, disagreements are adjudicated, and ambiguous,
severely occluded, or specialist-knowledge items are removed. This process, together with a subsequent rights and consent
review, retains 1,586 of 2,100 candidates, correcting 121 counts and replacing 79 images. Full sourcing, schema, gating, and
exclusion details are in \Cref{sec:collection-details}.

%% file: sec4_setup_and_models.tex
\section{Cross-Model Results}
\label{sec:eval}
\label{sec:setup}
\label{sec:metrics}

We use a paired evaluation protocol to compare frontier systems, then isolate the effects of model
scale and reasoning effort before testing alternative explanations for the benchmark gap.
Models receive the benchmark text verbatim, with no system prompt, exemplars, or output-format
instruction. Hosted
models use provider defaults, and open-weight models use their reference decoding settings.
Intervals are \SI{95}{\percent}, paired tests use exact McNemar, and the measured noise floor is
1.93 percentage points (\Cref{sec:variance,tab:noise}); media processing, parsing, decoding, and
per-run settings are detailed in \Cref{sec:impl}. Results in this section average three
independent repeats unless explicitly stated otherwise.

Answers are extracted \emph{position-first} ---
the last line containing an integer, a bolded value breaking ties --- because models state
intermediate quantities under labels such as ``total,'' which a label-first parser mis-scores; a
response containing no integer counts as wrong on accuracy and is excluded from MAE. Each per-family metric is calculated on the subset of images carrying the decoy of that family ($n=214$--$272$). An item with decoys from two families appears once in each
family denominator.

We use three complementary metrics:
\begin{enumerate}[leftmargin=*,itemsep=1pt,topsep=2pt]
  \item \textbf{Exact-match accuracy} ($A$): whether the extracted integer equals the true count;
  it measures success, not failure type.
  \item \textbf{Exact decoy-hit}: whether a wrong answer equals a pre-registered decoy, a conservative
  marker of a registered shortcut rather than an arbitrary miscount. No decoy equals its item's true
  count, so exact-match and decoy-hit are disjoint by construction.
  \item \textbf{Reasoning-alignment gap}:
  $\rgap=A_{\mathrm{cued}}-A_{\mathrm{neutral}}$, the paired exact-match gain from a matched
  alignment cue.
  This gap quantifies one component of visual-reasoning misalignment, not all reasoning or
  perceptual error.
\end{enumerate}

\subsection{Frontier Model Comparison}
\label{sec:baselines}
\label{sec:xmodel}

We first ask whether the same alignment failure appears across model families, rather than being an
artifact of one architecture, provider, or decoding stack.

\begin{figure}[h]
\centering
\safegraphic[width=\linewidth]{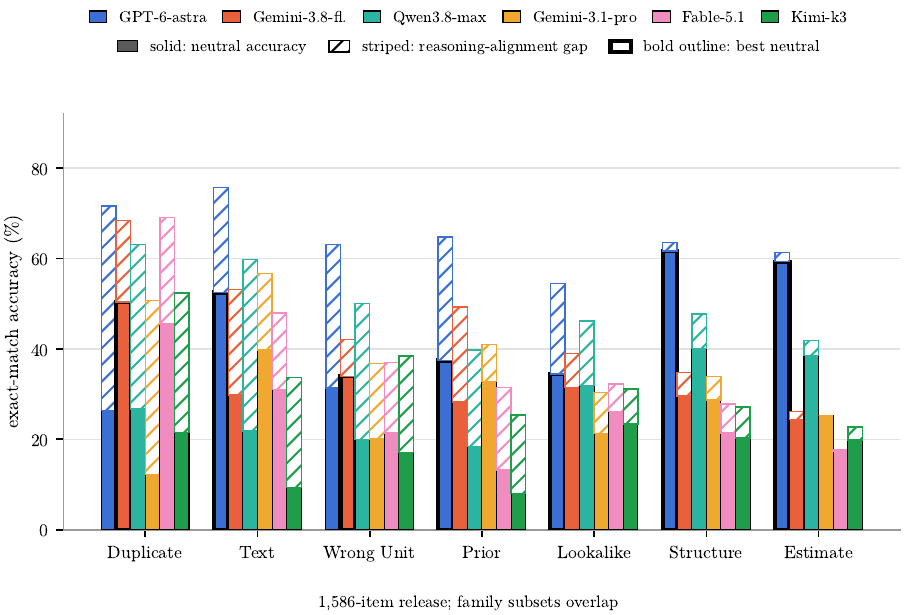}
\caption{\textbf{High capability does not imply a small reasoning-alignment gap.} Solid bars show
neutral accuracy and striped extensions show the gain from the matched alignment cue; bold outlines
mark the best neutral score in each family. GPT-6-astra leads the neutral score in five of the seven
families, and Gemini-3.8-flash in the other two, yet every model retains a large
reasoning-alignment gap in the five cue-responsive families. The smaller Structure and Estimate gaps instead identify
failures that guidance does not readily repair (\Cref{sec:gap}); exact values and decoy-hit rates are
in \Cref{tab:famacc,tab:decoy}.}
\label{fig:capture}
\end{figure}

\Cref{tab:models} summarizes six models on identical items, averaging three repeats.
Neutral accuracy remains below \SI{50}{\percent}; five models return a decoy more often than the
truth, and every $\rgap$ exceeds the 1.93-percentage-point noise floor. Accuracy, decoy-hit and gap do
not induce the same model ranking. These cross-system comparisons vary architecture, data and
training jointly and are descriptive; the controlled size ladder in \Cref{sec:scale} isolates scale.
\Cref{fig:capture} shows that family order also differs from aggregate capability.

\begin{table}[h]
\centering
\small
\begin{tabular}{@{}lcccccc@{}}
\toprule
& \multicolumn{3}{c}{\textbf{Accuracy}} & \multicolumn{2}{c}{\textbf{Decoy-hit}} & \\
\cmidrule(lr){2-4}\cmidrule(lr){5-6}
\textbf{Model} & \textbf{Neutral} & \textbf{Cued} & \textbf{multi-hop} & \textbf{Neutral} & \textbf{Cued} & \textbf{$\rgap$} \\
\midrule
GPT-6-astra            & \textbf{43.11} & \textbf{65.22} & \textbf{42.70} & 40.18 & \textbf{17.02} & \textbf{+22.11} \\
Gemini-3.8-flash       & 32.32 & 44.41 & 32.43 & \textbf{37.28} & 21.48 & +12.08 \\
Qwen3.8-max            & 28.21 & 49.94 & 32.79 & 50.74 & 25.51 & +21.73 \\
Gemini-3.1-pro-preview & 25.18 & 38.71 & 27.23 & 39.97 & 23.08 & +13.54 \\
Claude-Fable-5.1       & 25.05 & 37.98 & 27.69 & 45.00 & 27.51 & +12.93 \\
Kimi-k3                & 17.13 & 32.91 & 20.88 & 55.21 & 36.68 & +15.78 \\
\bottomrule
\end{tabular}
\caption{\textbf{The strongest neutral model also has the largest gap.} GPT-6-astra leads neutral
accuracy at \SI{43.11}{\percent} but gains another 22.11 percentage points when cued; Qwen3.8-max shows
a similarly large 21.73-point gap. Thus greater capability can improve what a model executes
under guidance without aligning what it selects unaided. The multi-hop condition uses the valid paired subset and
is not directly comparable with neutral or cued accuracy (\Cref{sec:impl}); bold marks column maxima.}
\label{tab:models}
\end{table}
\FloatBarrier

\subsection{Scale and Reasoning Effort}
\label{sec:scale}
Cross-model comparisons confound architecture and training, so we use controlled sweeps to
test whether greater capacity or inference-time reasoning closes the alignment gap.

\paragraph{Scaling widens the gap.}
Greater scale improves accuracy without improving visual decoy robustness. Across Qwen3.5
4B, 9B, and 27B, with image processing and decoding fixed, neutral accuracy rises from
14.08\% to 19.38\%, while cued accuracy rises from 22.91\% to 32.73\%.
The reasoning-alignment gap widens from 8.84 to 13.35 percentage points. The 4.52-point increase
has a paired 95\% CI of [2.77, 6.29], consistent with a path-selection failure:
the cue identifies a reasoning path
the model can execute but does not reliably choose unaided. Scale improves guided performance faster
than autonomous selection. The 9B-to-27B change is unresolved (CI [$-0.56$, 2.87]), so we report the
absolute gap and make no monotonic claim beyond the observed endpoints. Duplicate and Text retain
the largest gaps, and Structure and Estimate the smallest. This configuration-controlled result
supports the main analysis in \Cref{sec:gap}: capability can improve while the model remains
misaligned about which visual reasoning process to apply.

\begin{figure}[h]
\centering
\includegraphics[width=\linewidth]{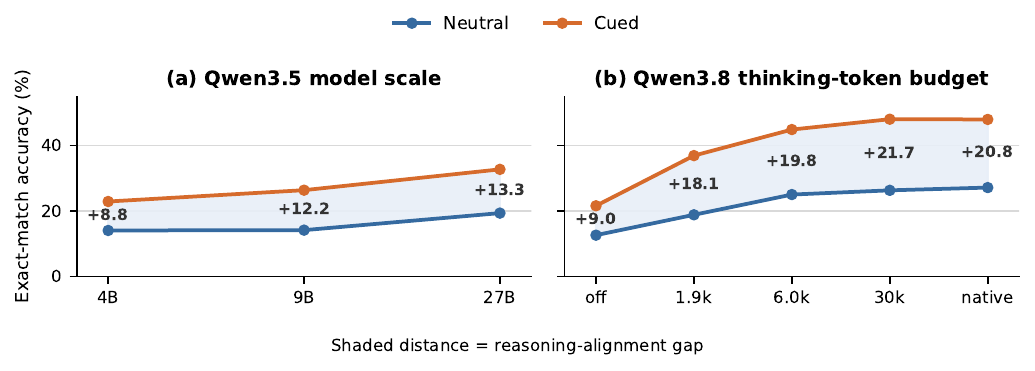}
\caption{\textbf{Scale and inference effort improve accuracy without closing the
reasoning-alignment gap.} The model-size sweep (left) and thinking-token sweep (right) compare
neutral and cued exact-match accuracy under controlled Qwen configurations. From 4B to 27B, cued
accuracy grows faster than neutral accuracy and $\rgap$ widens by 4.52 points; enabling native
thinking increases it by 11.84 points before it plateaus. Shaded distances and labels are $\rgap$
in percentage points; ``native'' denotes unbounded Qwen thinking. These sweeps retain their
original evaluation cohorts and have not been recomputed on the 1,586-item release.}
\label{fig:qwen35-scale}
\end{figure}

\paragraph{Scale concentrates errors on decoys.}
Neutral decoy-hit rises 43.76\%$\to$50.74\% from 4B to 27B, while other errors fall
42.17\%$\to$29.88\% (\Cref{fig:qwen35-error-composition}). Despite fewer mistakes overall,
a larger share of the remaining errors matches decoys: 50.92\%$\to$62.94\%
($+12.02$ points, paired 95\% CI [10.30, 13.74]). Thus both absolute decoy-hit and its share of
errors increase. Together with the widening reasoning-alignment gap, this shows that gains from scale do not
automatically align the model's default counting path with the intended task.

Image preprocessing and decoding are held fixed across sizes; full settings are in
\Cref{sec:qwen35-scale-config}.
\paragraph{A larger thinking budget widens the gap.} With self-hosted Qwen3.8 FlashNext fixed, the
shared effort setting resolves to Qwen's native control: thinking off; caps of 1,920, 6,048, or
30,000 tokens; or native unbounded thinking (\Cref{tab:qwen38-effort}). From off to native, neutral accuracy more than doubles
($12.65\to27.17\,\%$) while the gap grows from $+8.96$ to $+20.80$, peaking at $+21.71$ under the
30k-token cap (\Cref{fig:qwen35-scale}). This is a Qwen thinking-budget sweep, not a hosted-API
effort parameter. Full settings and run provenance are in \Cref{sec:qwen38-effort-config}.

\paragraph{Temperature does not explain the gap.} From $0.3$ to $1.0$,
neutral accuracy moves by at most $0.88$ points. At $T=1.5$ it collapses mainly because completions
stop terminating (\Cref{tab:temperature}).

\subsection{Validity Checks}
\label{sec:validity}
We test whether the headline effect is instead an artifact of edited images, wording ambiguity, or
adversarial selection. Each check asks whether the benchmark remains informative beyond its
construction choices.

\paragraph{Editing raises difficulty, not misalignment.} Most decoys must be
constructed rather than found, so \Cref{tab:provenance} separates commissioned, found and edited
images. The groups sample the taxonomy unevenly and are not interchangeable populations. Even after
matching on ground-truth count, edited-image accuracy is 8.4--22.8 percentage points lower than on
commissioned originals (all $p<0.001$), while found images remain close to originals. Decoy-hit is
also 21.6--24.8 percentage points higher on edited images.

The reasoning-alignment gap does not follow this difficulty shift: edited versus unedited gaps are
23.7 versus 20.3 percentage points for Qwen3.8-max ($p=0.15$), and
24.7 versus 20.3 percentage points for GPT-6-Astra ($p=0.43$). Editing changes absolute
difficulty, not the share of failure repaired by a cue. 

\paragraph{Model rankings persist despite adversariality.} Items were retained partly
because frontier models fail them (\Cref{sec:annotation}), so absolute accuracy is not a population
estimate of counting ability across a regular sample of images. Instead the image distribution is likely adversarial and out-of-distribution, especially with the addition of synthetic edits. We instead test whether \dataset{} preserves the model ordering of
standard counting benchmarks rather than measuring an unrelated out-of-distribution artifact.
Across nine models, rank correlation with CountQA is $\rho=0.98$ for neutral and $0.97$ for cued
accuracy; against FSC147 it is $0.83$ and $0.94$ (\Cref{fig:vs-counting,app:comparison_to_countqa_fsc}). \dataset{}
therefore ranks models as established counting benchmarks do, rather than measuring an unrelated
out-of-distribution skill, while still separating the failure mechanisms those benchmarks aggregate
away.

\paragraph{Wording exposes unstable alignment.}
The neutral phrase \emph{are there} admits an in-image or in-the-world reading. Replacing only that
phrase with \emph{are visible} changes all but six prompts without adding answer-relevant
information. On one Gemini-3.8-flash rollout, accuracy rises from \SI{32.89}{\percent} to
\SI{39.58}{\percent}; 246 answers are repaired and 135 broken (exact McNemar
$p=1.39\times10^{-8}$). The effect is not uniform: Text and Prior improve the most, while Estimate
degrades (\Cref{tab:global-wording-ablation}). This is not a pure clarification: \emph{visible} can
change which instances a human considers members of the counted set. The large swing from a minor
surface change, therefore, exposes unstable alignment between the model's default interpretation and
the intended object set. This motivates testing whether a substantially richer, real-life multi-hop
prompt resolves the same misalignment (\Cref{sec:mh}).

%% file: sec5_main_findings.tex
\section{Understanding Visual-Reasoning Misalignment}
\label{sec:results}

We now use the paired design to demonstrate that our reasoning alignment gap is a core source of counting failures and provide evidence that this gap persists under multiple experimental setups and points to implicit model misalignment. We first test which errors a
matched cue repairs, then ask whether agentic tools or realistic context can provide the same
alignment without naming the trap.

\subsection{Reasoning Alignment Gap}
\label{sec:gap}

We begin by testing whether the alignment cue selectively repairs the pre-registered failure mode,
rather than merely eliciting more effort or broadly improving accuracy.

\paragraph{Matched cues repair decoy-aligned errors.}
No model exceeds \SI{50}{\percent} neutral accuracy, and five of six return the planted decoy more
often than the truth. Cuing the trap repairs 291--417 items while breaking only
59--126 (exact McNemar, $p<10^{-20}$ for every model), and releases
\SIrange{43.4}{65.8}{\percent} of exact decoy hits. This paired, directional effect operationalizes
visual-reasoning misalignment rather than merely recording another accuracy difference; full results are in
\Cref{tab:main}.

\paragraph{Misalignment is mechanism-dependent.}
Five families gain 10.78--32.11 percentage points from the cue, whereas Structure gains
5.51 points and Estimate 1.74 points (\Cref{tab:decoy,tab:famacc}). No model gains more than
3.35 points on Estimate, consistent with a perceptual or enumeration bottleneck that guidance
alone does not resolve. A small cue gain does not by itself distinguish perception from other
reasoning limitations. This split holds
across all six models, but the ordering within the five large-gap families does not; aggregate
accuracy therefore hides large, model-specific failure modes.

\paragraph{Decoy share separates shortcut from miscount.} Among the five large-gap families,
\SIrange{61.5}{81.4}{\percent} of errors equal the planted decoy (\Cref{tab:mechanism}). Estimate is
only \SI{21.5}{\percent}: most failures are ordinary miscounts. Structure is intermediate at
\SI{49.5}{\percent}, mixing lattice shortcuts with miscounting. Its small gap therefore contains a
repairable shortcut component, whereas Estimate is consistent with a perceptual or enumeration bottleneck.

\paragraph{The reasoning-alignment gap depends on trap-specific cues.}
\label{sec:robustness}
To separate relevant information from generic effort, we compare neutral, generic ``count
carefully,'' mismatched-family, and matched-family prompts on Gemini-3.8-flash and GPT-6-astra, with
three repeats each (\Cref{tab:cuecontrol}).

\begin{table}[!htbp]
\centering
\scriptsize
\renewcommand{\arraystretch}{0.92}
\begin{tabular}{@{}lccccc@{}}
\toprule
\textbf{Condition} & \textbf{Acc.} & \textbf{vs.\ neutral} & \textbf{\% of matched} & \textbf{Decoy-hit} & \textbf{Compl.\ tok.} \\
\midrule
\multicolumn{6}{@{}l}{\emph{Gemini-3.8-flash}} \\
Neutral      & 32.32 & ---     & ---  & 37.28 & 5297 \\
Generic cue  & 34.62 & $+$2.29 & \phantom{0}19 & 33.54 & 6258 \\
Mismatched cue & 34.70 & $+$2.37 & \phantom{0}20 & 32.53 & 5407 \\
Matched cue  & \textbf{44.41} & $\mathbf{+12.08}$ & 100 & \textbf{21.48} & 6263 \\
\addlinespace[3pt]
\multicolumn{6}{@{}l}{\emph{GPT-6-astra}} \\
Neutral      & 43.11 & ---     & ---  & 40.18 & 2581 \\
Generic cue  & 46.17 & $+$3.07 & \phantom{0}14 & 35.62 & 3267 \\
Mismatched cue & 44.79 & $+$1.68 & \phantom{00}8 & 34.05 & 2957 \\
Matched cue  & \textbf{65.22} & $\mathbf{+22.11}$ & 100 & \textbf{17.02} & 3233 \\
\bottomrule
\end{tabular}
\caption{\textbf{Generic and mismatched cues recover at most 20\% of the matched gain:} improvement depends
on naming the actual trap, not merely eliciting more effort. Tokens include reasoning.}
\label{tab:cuecontrol}
\end{table}

Thus at least \SI{80}{\percent} of the gain depends on cuing the trap actually present. The matched cue separates from both controls
at $p<10^{-12}$, while the controls do not separate from each other. Completion length cannot
explain the result: generic and matched prompts use nearly identical budgets on Gemini, and generic
uses more on GPT-6 while producing one eighth of the gain.

The exceptions sharpen the mechanism (\Cref{tab:cuefamily}). Duplicate gains substantially only when its own trap is cued;
Estimate barely moves under any cue; and Text gains about 11 percentage points even from generic or
mismatched instructions, suggesting that any request to count can displace reading the printed
numeral. The \SI{80}{\percent} bound therefore describes the aggregate, not every family.

\subsection{Agentic Harnesses and Tools}
\label{sec:whentools}

We next test whether additional computation or visual instrumentation can replace explicit
guidance, separating gains in task accuracy from changes in the reasoning-alignment gap.

\paragraph{Tools can improve accuracy without improving alignment.}
A matched cue identifies the trap; a tool measures or computes. We define
$\rgapp=\rgap_{\text{tool}}-\rgap_{\text{base}}$: negative values narrow the reasoning-alignment
gap, while positive values widen it. This separates task improvement from alignment. For example,
GPT-6 with CI raises neutral accuracy from \SI{27.31}{\percent} to \SI{35.72}{\percent} and cued
accuracy from \SI{46.65}{\percent} to \SI{57.65}{\percent}, yet widens the gap by
2.59 percentage points. Generic CI usually follows this pattern: it adds computation and deliberation,
but does not identify the intended visual counting process. This aggregate trend is not universal across families:
GPT-6 CI narrows the gap for Wrong Unit by 0.8 percentage points, and Opus CI narrows it for Duplicate
by 1.7 points (\Cref{tab:toolfamily}). These exceptions suggest that generic execution can
occasionally supply evidence useful for a particular failure mode, but does not do so reliably.
Note that a positive aggregate $\rgapp$ does not show that tools harm alignment: provider-native
CI also changes the length and structure of the reasoning process, allowing the cued arm to benefit
more even as neutral accuracy rises.

\begin{table}[!htbp]
\centering
\small
\setlength{\tabcolsep}{7pt}
\renewcommand{\arraystretch}{1.12}
\begin{tabular}{@{}llrrr@{}}
\toprule
\textbf{System} & \textbf{Comparison} & \textbf{$\rgap_{\rm base}$} &
\textbf{$\rgap_{\rm tool}$} & \textbf{$\rgapp$} \\
\midrule
Qwen3.8-Flash-Next & bare $\to +$CI$^*$       & 19.46 & 26.63 & \gapwidens{+7.17} \\
Qwen3.8-Flash-Next & $+$CI $\to +$CI$+$SAM3 & 26.63 & 22.11 & \gapnarrows{-4.52} \\
GPT-6-sol          & no tools $\to +$CI        & 19.34 & 21.93 & \gapwidens{+2.59} \\
Claude-Opus-5.5    & no tools $\to +$CI        & 11.97 & 14.45 & \gapwidens{+2.48} \\
\bottomrule
\end{tabular}
\caption{\textbf{General agentic harnesses widen the gap; targeted SAM3 narrows it.}
Change on the full release (points):
\textcolor{gapgreen}{\textbf{green}} narrows the gap; \textcolor{gapred}{\textbf{red}} widens it.
CI is provider-native execution; $^*$ also changes the agent. Details:
\Cref{tab:toolraw}.}
\label{tab:tooluse}
\end{table}

\paragraph{Targeted visual tools narrow the gap.}
The CI comparisons also induce longer, multi-turn reasoning, so they do not show that computation itself harms alignment. The cleaner fixed-agent
CI$\to$CI+SAM3 contrast raises neutral accuracy by 3.61 percentage points and narrows the gap by
4.52 points. SAM3 narrows all seven families, most strongly Estimate
($-10.8$ points), Structure ($-8.6$ points), and Text
($-5.4$ points), whereas CI alone narrows only Wrong Unit for GPT-6 and Duplicate for Opus.
The distinction is therefore not tools versus no tools, but whether the tool supplies evidence for
the family-specific visual bottleneck, an effect that itself depends on the agentic
surface the tool is attached to (\Cref{tab:toolfamily,fig:d2r,sec:secondary-agent}).

\subsection{Contextual Grounding}
\label{sec:mh}

\begin{figure}[t]
\centering
\includegraphics[width=\linewidth]{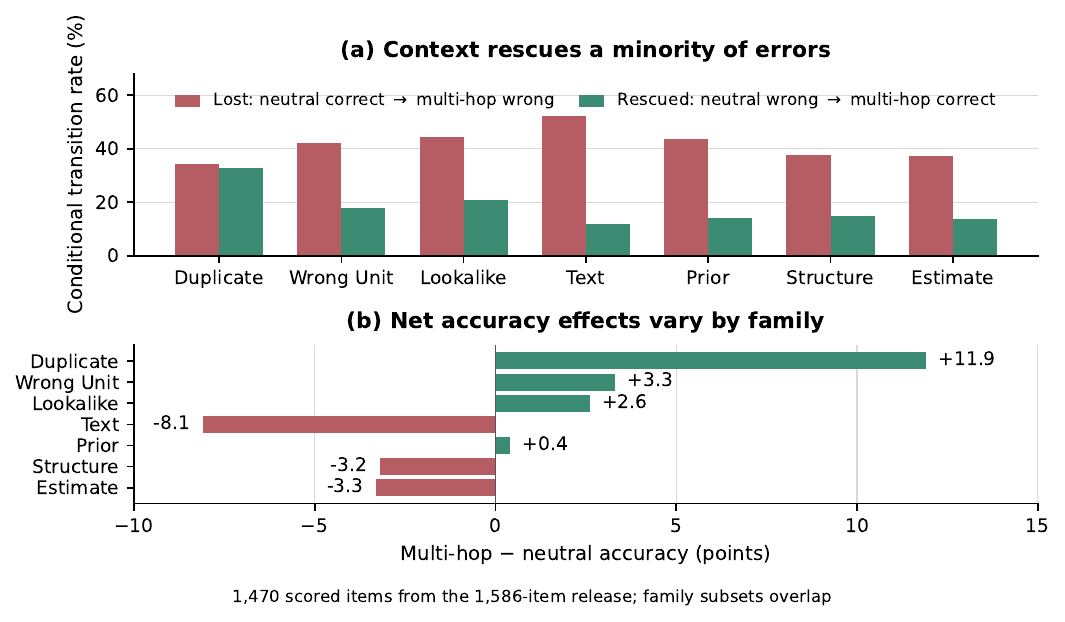}
\caption{\textbf{Scenario grounding resolves some ambiguity but not the full alignment gap.}
For each family, the top panel reports losses among neutral-correct items and rescues among
neutral-wrong items; the bottom panel shows the resulting change in accuracy. Context rescues only
11.9--32.7\% of neutral failures and helps most on Duplicate, where the scenario clarifies what
counts as one usable object. Text, Structure, and Estimate decline, showing that added context does
not generally replace a targeted alignment cue. ``Lost'' denotes neutral-correct $\to$ multi-hop-wrong;
``rescued'' denotes neutral-wrong $\to$ multi-hop-correct. Exact transition counts remain in
\Cref{tab:mhdecomp}.}
\label{fig:mh-context}
\end{figure}

The multi-hop condition tests whether neutral failures arise because a short question does not say
enough about what should count. It embeds the same count in a realistic scenario without cuing the
decoy mechanism, making it a contextual control rather than a ladder of reasoning difficulty. A
bare ``How many CDs are there?'' leaves open whether a reflected CD counts; ``I want to give each
friend 2 CDs'' does not. If vagueness caused the failures, context should fix most of
them. Yet it does not: of the items a model miscounted under the neutral prompt, only
\SIrange{11.9}{32.7}{\percent} per family are answered correctly once the purpose is given, and the
rest still fail (\Cref{tab:mhdecomp}). What context does recover is family-dependent: it helps where
it settles what counts as one object, recovering an average of \SI{23.7}{\percent} of miscounted items in the three
unit-settling families (Duplicate, Wrong Unit, Lookalike) against an average of \SI{13.4}{\percent} in the other
four (\Cref{fig:mh-context}).

The family directions clarify what context does and does not repair. Duplicate gains most because a
real-life purpose can establish whether a reflection or depiction is a second usable object. Text
instead loses the most: adding prose and arithmetic does not directly suppress the answer-like
numeral, and creates another route to error. Structure and Estimate also remain negative, consistent
with failures that richer semantics cannot resolve. These per-family directions hold within each
model as well as in aggregate (\Cref{tab:mhfamily}).

The initial visual count remains the bottleneck. Across models, multi-hop accuracy is
24.1--49.3 percentage points higher when the paired neutral count is correct
(\Cref{tab:mhprop}). Context can occasionally settle the unit, but downstream reasoning rarely
compensates for a wrong count already extracted from the image.

Aggregate accuracy barely moves because the extra arithmetic loses
\SIrange{34.4}{52.1}{\percent} of neutral-correct items across decoy families (\Cref{tab:mhdecomp}), whereas
cuing the trap raises original-count accuracy by 12.1--22.1 percentage points across models without changing
the task (\Cref{tab:models}). Missing
context is therefore not the main explanation; with the scale, effort, and tool results, this makes
visual-reasoning misalignment a measurable but incomplete reasoning failure. Remaining errors can
still reflect perception or other reasoning limitations.


%% file: appendix.tex
\appendix

\section{Implementation Details}
\label{sec:impl}
\Cref{sec:setup} states the inference configuration; this section records the parts of it a
re-implementation would have to reproduce exactly.

\paragraph{Media preparation.} The delivery contains HEIC, WebP, GIF and SVG files, 47 files with no
extension, and images up to \SI{30.7}{\mega\byte}. We convert all 2502 to JPEG at quality 92, rasterising
SVG first, with the longest side capped at 2048 pixels --- the resolution the receiving API downsamples
to anyway, so the cap is approximately lossless with respect to what the model observes. No image failed.

\paragraph{Answer extraction.} Responses are parsed position-first: the last line containing an integer
supplies the answer, a bolded value breaking ties. This matters --- an earlier label-first parser
preferring a line reading ``total'' or ``answer'' mis-scored genuine outputs, because models state
intermediate quantities under exactly those labels. The extractor carries a regression test.

\paragraph{Exclusions.} Every item ships a primary image, so none is excluded on that ground. The
803 items without a secondary image have self-contained multi-hop questions and are evaluated
single-image rather than dropped. The multi-hop condition is scored on 1,470 items: the 116 whose counts were corrected in quality
control (\Cref{sec:annotation}) kept a derived answer computed from the old count, so their
multi-hop ground truth is known to be wrong.

\paragraph{Configuration.} Gemini-3.8-flash, reasoning effort \texttt{medium}, output limit 16384
tokens, 128-way concurrency, five attempts with exponential backoff. One pass over the frozen
release in all three conditions is \num{4980} calls, \num{6424061} prompt tokens and
\num{29136062} completion tokens (of which \num{28822703} are reasoning); the headline figures in
\Cref{tab:models} average three such passes per condition. Every call succeeded and every response
terminated normally.

\paragraph{Decoding.} Hosted models receive no sampling override and use provider defaults. Locally
served open-weight models use temperature 0.6, top-$p$ 0.95, and top-$k$ 20. Generation limits span
65,536--128,000 tokens and do not bind; reasoning effort is fixed within model and condition except
in the controlled sweep of \Cref{sec:scale}.

\clearpage
\subsection{Release Composition}
This expands the taxonomy and domain summary used in \Cref{sec:decoycount}.
\begin{table}[h]
\centering
\small
\renewcommand{\arraystretch}{0.95}
\begin{tabular}{@{}lp{0.37\linewidth}p{0.39\linewidth}@{}}
\toprule
\textbf{Family} & \textbf{How the decoy is induced} & \textbf{What the matched cue tests} \\
\midrule
Duplicate & One physical instance appears more than once through a reflection, screen, or depiction.
          & Whether the model distinguishes entities from their representations. \\
Text & A printed numeral offers an answer-like shortcut.
     & Whether the model counts visible objects rather than transcribing text. \\
Wrong Unit & Groups, containers, or subparts invite counting at the wrong granularity.
           & Whether the model resolves the unit requested by the question. \\
Prior & A familiar arrangement suggests a canonical count unsupported by the image.
      & Whether visual evidence overrides a memorized or expected answer. \\
Lookalike & Target-like distractors blur the category boundary.
          & Whether the model maintains the intended object class while counting. \\
Structure & A regular row, column, or lattice invites an incorrect structural shortcut.
          & Whether the model verifies the arrangement instead of extrapolating a pattern. \\
Estimate & A dense or irregular field encourages approximate rather than exact enumeration.
         & Whether guidance can overcome a perceptual enumeration bottleneck. \\
\bottomrule
\end{tabular}
\caption{\textbf{The taxonomy localizes where counting can fail.} Each family pre-registers both an
answer-like decoy and a targeted intervention, allowing the paired evaluation to distinguish
the reasoning-alignment gap from residual perceptual and other reasoning failures.}
\label{tab:taxonomy}
\end{table}
\FloatBarrier

\begin{table}[h]
\centering
\small
\begin{tabular}{@{}lrrr@{}}
\toprule
\textbf{Decoy family} & \textbf{Decoys} & \textbf{Items} & \textbf{Scene domain} \hfill \textbf{Items} \\
\midrule
Estimate    & 476 & 249 & Everyday \hfill 318 \\
Lookalike   & 346 & 272 & Public \& outdoor \hfill 257 \\
Structure   & 290 & 246 & Documents, screens \& artwork \hfill 234 \\
Wrong Unit  & 260 & 221 & Charts, maps \& scientific \hfill 222 \\
Prior       & 257 & 243 & Observation \hfill 191 \\
Duplicate   & 256 & 236 & Retail \& product \hfill 186 \\
Text        & 223 & 214 & Commercial \& industrial \hfill 178 \\
\midrule
\textbf{Total} & \textbf{2,108} & & \hfill \textbf{1,586} \\
\bottomrule
\end{tabular}
\caption{Release composition ($n=1,586$). Every item has at least one decoy; 376 have multiple,
yielding 2,108 registered decoys. Items contribute once to every family represented by their decoys,
whereas scene domains are mutually exclusive.}
\label{tab:families}
\end{table}
\FloatBarrier

The scene domains are Everyday; Public \& outdoor; Documents, screens \& artwork; Charts, maps \&
scientific; Retail \& product; Observation; and Commercial \& industrial. They are tracked separately
from mechanism labels so that a family does not collapse onto one kind of scene.

The 2,108 assignments across 1,586 items also make the taxonomy explicitly multi-label: one scene can
invite several incorrect counting processes at once. The families therefore identify failure
mechanisms rather than disjoint visual categories. Their distribution across seven scene domains
reduces the risk that a mechanism score is merely a proxy for one recurring setting.

\subsection{Collection and Annotation Details}
\label{sec:collection-details}
Images come from commissioned photography, Wikimedia Commons, Openverse, CC0/CC BY Flickr material,
and government archives. Images are edited or generated when a mechanism cannot be found naturally.
The source delivery records provenance and licence fields, although these are not carried into the
released parquet.

Each item records a ground-truth count, up to five decoys, task type, scene description, three prompts,
an optional second image, and source metadata. Counts and decoys are written before model exposure.
The multi-hop task is retained only when the decoy and true count lead to different final answers.

Candidates pass a model gate using Claude Opus 5, GPT-5.6 Sol, and Gemini 3.6 Flash: all three must
miss the neutral answer and at least one must return a registered decoy exactly. Independent reviewers
then verify that the target is countable, visually grounded, and unambiguous. Blind counters assess
neutral and contextual tasks separately; disagreements receive a final adjudication.

Quality control corrected 121 counts, replaced 79 images, and removed ambiguous or specialist items,
including severe occlusion, distant objects indistinguishable from background, partial objects at
image boundaries, and questions requiring non-generalist notation knowledge. The main experimental
contract is summarized in \Cref{sec:collection}.

\subsection{Paired Repair Results}
\begin{table}[h]
\centering
\small
\setlength{\tabcolsep}{4pt}
\begin{tabular}{@{}lccccccc@{}}
\toprule
& \multicolumn{2}{c}{\textbf{Neutral}} & \textbf{Cued} & \multicolumn{3}{c}{\textbf{Paired neutral $\to$ cued}} & \textbf{Bites} \\
\cmidrule(lr){2-3}\cmidrule(lr){4-4}\cmidrule(lr){5-7}
\textbf{Model} & \textbf{Acc.} & \textbf{Decoy-hit} & \textbf{Acc.} & \textbf{Repaired} & \textbf{Broken} & \textbf{$p$} & \textbf{released} \\
\midrule
GPT-6-astra       & \textbf{43.11} & 40.18 & \textbf{65.22} & 417 & \phantom{0}66 & $p<10^{-20}$ & \textbf{65.8} \\
Gemini-3.8-flash  & 32.32 & 37.28 & 44.41 & 299 & 107 & $p<10^{-20}$ & 55.6 \\
Qwen3.8-max       & 28.21 & 50.74 & 49.94 & 404 & \phantom{0}59 & $p<10^{-20}$ & 55.8 \\
Gemini-3.1-pro    & 25.18 & 39.97 & 38.71 & 341 & 126 & $p<10^{-20}$ & 61.4 \\
Claude-Fable-5.1  & 25.05 & 45.00 & 37.98 & 291 & \phantom{0}86 & $p<10^{-20}$ & 50.5 \\
Kimi-k3           & 17.13 & \textbf{55.21} & 32.91 & 323 & \phantom{0}72 & $p<10^{-20}$ & 43.4 \\
\bottomrule
\end{tabular}
\caption{Matched cues repair 2.7--6.8$\times$ more items than they break and remove 43--66\% of
neutral decoy hits. \emph{Repaired} and \emph{broken} are paired flips; $p$ is exact McNemar.}
\label{tab:main}
\end{table}
\FloatBarrier

The paired direction is consistent across models: a matched cue repairs between 291 and 417 answers,
whereas only 59--126 correct answers become wrong. This rules out a generic prompt-perturbation
explanation for the aggregate gains. The variation in decoy bites released further shows that
alignment gap and decoy capture are related but distinct: a cue may redirect an incorrect process
without every repaired response having first landed on a registered decoy.

\subsection{Qwen3.5 Scale-Study Configuration}
\label{sec:qwen35-scale-config}
Self-hosted vLLM, no tools, 5MP images, native thinking without
a separate reasoning budget; temperature 1, top-$p$ 0.95, top-$k$ 20, min-$p$ 0, presence penalty
1.5, repetition penalty 1, and a 131,072-token cap that binds on some responses. These settings
differ from \Cref{sec:setup}. Release alignment removes 29 items and corrects 11 decoy annotations;
retained inputs and gold answers are unchanged. Pointwise intervals use 10,000 paired
primary-image-cluster bootstrap draws (seed 20260924), averaging repeats within items;
conditional decoy-hit is a ratio of summed counts.

\subsection{Qwen3.8 FlashNext Thinking-Budget Sweep}
\label{sec:qwen38-effort-config}
This experiment uses self-hosted \texttt{Qwen/Qwen3.8-Flash-Next} through vLLM, not a hosted API's
reasoning-effort parameter. The shared \texttt{--reasoning-effort} interface resolves to Qwen's
native controls: \texttt{none} disables thinking; \texttt{low}, \texttt{medium}, and \texttt{high}
set thinking-token caps of 1,920, 6,048, and 30,000; and \texttt{native} enables thinking without a
separate cap. All conditions use 5MP images, three repeats, temperature 1, top-$p$ 0.95, top-$k$ 20,
and a 4,096-token response cap.

\begin{table}[h]
\centering \small
\setlength{\tabcolsep}{4pt}
\begin{tabular}{lrrrrr}
\toprule
\textbf{Thinking} & \textbf{Neutral run} & \textbf{Cued run} & \textbf{Neutral} & \textbf{Cued} & $\boldsymbol{\rgap}$ \\
\midrule
Off       & 48345 & 49270 & 12.65 & 21.61 &  8.96 \\
1,920     & 48347 & 49271 & 18.86 & 36.93 & 18.07 \\
6,048     & 48348 & 49272 & 25.04 & 44.88 & 19.84 \\
30,000    & 48343 & 49269 & 26.33 & 48.03 & \textbf{21.71} \\
Native    & 48342 & 49268 & \textbf{27.17} & 47.97 & 20.80 \\
\bottomrule
\end{tabular}
\caption{Qwen3.8 FlashNext effort sweep. Accuracy is in percent and $\rgap$ in percentage points, averaged over three
rollouts per item. ``Native'' means unbounded Qwen thinking, so the gap peaks at the 30k cap rather
than at the endpoint. Run IDs identify the complete paired artifacts in \texttt{s3://elorian-eval}.}
\label{tab:qwen38-effort}
\end{table}
\FloatBarrier

\begin{figure}[h]
\centering
\includegraphics[width=\linewidth]{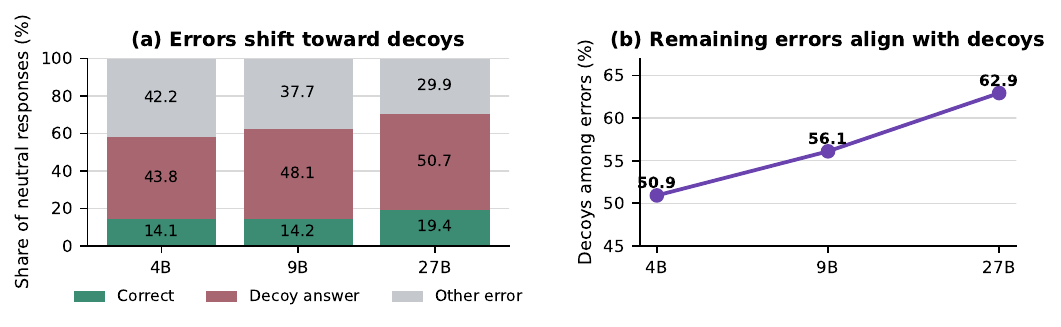}
\caption{\textbf{Scale reduces miscellaneous errors faster than decoy-aligned errors.} The stacked
bars (left) partition neutral responses into correct, exact-decoy, and other-error outcomes; the
line (right) reports the decoy share among all remaining errors. From 4B to 27B, accuracy rises by
5.30 points and other errors fall by 12.29 points, yet exact-decoy answers rise by 6.98 points.
Accordingly, the decoy share of errors increases by 12.02 points (paired 95\% CI [10.30, 13.74]).
These are the original scale-study measurements, not a rescore on the 1,586-item release.}
\label{fig:qwen35-error-composition}
\end{figure}
\FloatBarrier

Scaling removes miscellaneous errors faster than it removes registered shortcuts. The larger model
is therefore more accurate overall, but a greater fraction of its residual errors follows a specific,
predictable alternative counting process. This is the error-composition counterpart of the widening
reasoning-alignment gap: capability growth does not ensure that the model selects the intended process.

\begin{table}[h]
\centering \small
\setlength{\tabcolsep}{6pt}
\begin{tabular}{lcc}
\toprule
\textbf{Temperature} & \textbf{Neutral acc.} & $\Delta$ vs.\ 1.0 \\
\midrule
0.3 & \textbf{28.05} & $+0.88$ \\
0.6 & \textbf{28.05} & $+0.88$ \\
1.0 & 27.17 & --- \\
1.5 & 11.93 & $-15.24$ \\
\bottomrule
\end{tabular}
\caption{Neutral exact-match accuracy for Qwen3.8 FlashNext at native unbounded thinking, averaged
over three repeats (runs 48349, 48350, 48342, and 48351). Temperature is flat through 1.0; at 1.5,
\SI{12.73}{\percent} of rollouts return no integer.}
\label{tab:temperature}
\end{table}
\FloatBarrier

The near-identical results from temperature 0.3 through 1.0 make sampling temperature an unlikely
explanation for the scale trend. The collapse at 1.5 is qualitatively different: missing integer
answers indicate generation instability rather than a change in counting alignment.

\subsection{Provenance Control}
This is the full provenance breakdown behind the control in \Cref{sec:validity}.
\begin{table}[h]
\centering
\small
\begin{tabular}{@{}lrcccccc@{}}
\toprule
& & \multicolumn{3}{c}{\textbf{Qwen3.8-max}} & \multicolumn{3}{c}{\textbf{GPT-6-astra}} \\
\cmidrule(lr){3-5}\cmidrule(lr){6-8}
\textbf{Provenance} & \textbf{$n$} & \textbf{Neutral} & \textbf{Cued} & \textbf{$\rgap$} & \textbf{Neutral} & \textbf{Cued} & \textbf{$\rgap$} \\
\midrule
Original photography    & 402 & \textbf{41.4} / 37.6 & \textbf{55.5} / 18.9 & $+14.1$ & \textbf{54.9} / 26.9 & \textbf{69.3} / 11.9 & $+14.4$ \\
Public domain, as found & 527 & 30.5 / 46.2 & 55.6 / 19.9 & $+25.1$ & 41.6 / 39.9 & 66.3 / 13.9 & $+24.7$ \\
Edited                  & 657 & 18.3 / 62.4 & 42.0 / 34.1 & $+23.7$ & 37.1 / 48.5 & 61.8 / 22.6 & $+24.7$ \\
\midrule
\multicolumn{8}{@{}l}{\emph{Count-matched against original photography (same $y$; $\Delta$ accuracy)}} \\
Public domain, as found & 292 & $-7.2$ & $-2.4$ & & $-7.8$ & $-6.1$ & \\
Edited                  & 334 & $\mathbf{-22.8}$ & $\mathbf{-13.4}$ & & $\mathbf{-17.1}$ & $\mathbf{-8.4}$ & \\
\bottomrule
\end{tabular}
\caption{Accuracy / decoy-hit by provenance (\si{\percent}). Bottom rows count-match against
commissioned originals; differences and $\rgap$ are in percentage points. Editing changes accuracy but not $\rgap$.}
\label{tab:provenance}
\end{table}
\FloatBarrier

Edited images are harder in absolute terms, but their reasoning-alignment gaps are comparable to those of
unmodified public images. The phenomenon is thus not confined to synthetic or deliberately edited
scenes. Because provenance groups also differ in taxonomy and image content, this control supports
robustness rather than a causal claim that editing has no effect.

\clearpage
\subsection{Family-Level Alignment Results}
These tables provide the model- and family-level values summarized in \Cref{sec:gap}.
\begin{table}[h]
\centering
\tiny
\setlength{\tabcolsep}{2.5pt}
\begin{tabular}{@{}lcccccccccccc@{}}
\toprule
& \multicolumn{2}{c}{\textbf{GPT-6-astra}} & \multicolumn{2}{c}{\textbf{Gemini-3.8-fl.}} & \multicolumn{2}{c}{\textbf{Qwen3.8-max}} & \multicolumn{2}{c}{\textbf{Gemini-3.1-pro}} & \multicolumn{2}{c}{\textbf{Fable-5.1}} & \multicolumn{2}{c}{\textbf{Kimi-k3}} \\
\cmidrule(lr){2-3}\cmidrule(lr){4-5}\cmidrule(lr){6-7}\cmidrule(lr){8-9}\cmidrule(lr){10-11}\cmidrule(lr){12-13}
\textbf{Family} & \textbf{N.} & \textbf{C.} & \textbf{N.} & \textbf{C.} & \textbf{N.} & \textbf{C.} & \textbf{N.} & \textbf{C.} & \textbf{N.} & \textbf{C.} & \textbf{N.} & \textbf{C.} \\
\midrule
Duplicate  & 61.30 & 17.51 & 36.58 & 17.66 & 61.86 & 24.29 & 67.51 & 31.21 & 40.11 & 16.67 & 61.16 & 30.51 \\
Wrong Unit & 51.73 & 20.66 & 38.61 & 23.08 & 58.97 & 27.15 & 44.04 & 29.56 & 51.73 & 34.99 & 54.45 & 34.84 \\
Prior      & 48.42 & 22.09 & 50.07 & 27.16 & 68.45 & 39.92 & 39.92 & 22.09 & 74.07 & 47.60 & 74.35 & 52.40 \\
Lookalike  & 47.18 & 22.55 & 36.76 & 20.71 & 48.16 & 26.84 & 45.96 & 30.39 & 39.71 & 28.31 & 46.81 & 32.97 \\
Text       & 39.56 & 12.31 & 56.39 & 27.88 & 71.96 & 25.70 & 36.92 & 13.55 & 54.21 & 25.86 & 84.27 & 54.98 \\
Structure  & 21.41 & 17.21 & 27.91 & 22.76 & 32.66 & 24.25 & 30.89 & 24.39 & 37.26 & 28.32 & 45.66 & 36.99 \\
Estimate   & 8.84 & 7.63 & 15.39 & 12.72 & 15.13 & 11.78 & 10.04 & 8.43 & 18.88 & 14.46 & 22.09 & 15.66 \\
\bottomrule
\end{tabular}
\caption{Per-family neutral/cued decoy-hit, six models and three repeats (\si{\percent}). Bold
marks the largest neutral-to-cued reduction in each family.}
\label{tab:decoy}
\end{table}

\begin{table}[h]
\centering
\small
\setlength{\tabcolsep}{4pt}
\begin{tabular}{@{}lrccccccc@{}}
\toprule
\textbf{Family} & \textbf{$n$} & \textbf{GPT-6} & \textbf{G-3.8-fl.} & \textbf{Qwen-max} & \textbf{G-3.1-pro} & \textbf{Fable-5.1} & \textbf{Kimi-k3} & \textbf{Mean} \\
\midrule
Duplicate  & 236 & +45.20 & +17.94 & +36.44 & +38.56 & +23.59 & +30.93 & \textbf{+32.11} \\
Text       & 214 & +23.21 & +23.36 & +38.01 & +16.98 & +16.98 & +24.30 & \textbf{+23.81} \\
Wrong Unit & 221 & +31.67 & \phantom{0}+8.14 & +30.17 & +16.74 & +15.54 & +21.42 & \textbf{+20.61} \\
Prior      & 243 & +27.30 & +20.99 & +21.40 & \phantom{0}+8.23 & +18.24 & +17.42 & \textbf{+18.93} \\
Lookalike  & 272 & +19.98 & \phantom{0}+7.48 & +14.34 & \phantom{0}+9.07 & \phantom{0}+6.13 & \phantom{0}+7.72 & \textbf{+10.78} \\
Structure  & 246 & \phantom{0}+1.90 & \phantom{0}+5.28 & \phantom{0}+7.59 & \phantom{0}+5.15 & \phantom{0}+6.37 & \phantom{0}+6.78 & \phantom{0}\textbf{+5.51} \\
Estimate   & 249 & \phantom{0}+2.01 & \phantom{0}+1.74 & \phantom{0}+3.35 & \phantom{0}+0.00 & \phantom{0}+0.40 & \phantom{0}+2.95 & \phantom{0}\textbf{+1.74} \\
\bottomrule
\end{tabular}
\caption{Per-family $\rgap$ (percentage points). An item with decoys from multiple families contributes
once to each applicable family; multiple decoys from the same family do not duplicate the item.
Bold marks the cross-model mean used to order families.}
\label{tab:famacc}
\end{table}

\begin{table}[h]
\centering
\small
\begin{tabular}{@{}lrrrr@{}}
\toprule
\textbf{Family} & \textbf{Neutral} & \textbf{Cued} & \textbf{Gap} & \textbf{Errors on decoy} \\
\midrule
Text       & 30.7 & 54.5 & $+$23.8 & \textbf{81.4} \\
Duplicate  & 30.4 & 62.5 & $\mathbf{+32.1}$ & 78.3 \\
Prior      & 23.0 & 41.9 & $+$18.9 & 76.1 \\
Wrong Unit & 24.0 & 44.6 & $+$20.6 & 65.7 \\
Lookalike  & 28.1 & 38.9 & $+$10.8 & 61.5 \\
Structure  & 33.6 & 39.1 & \phantom{0}$+$5.5 & 49.5 \\
Estimate   & 30.8 & 32.5 & \phantom{0}$\mathbf{+1.7}$ & \textbf{21.5} \\
\bottomrule
\end{tabular}
\caption{Outcomes pooled over six models and three repeats (\si{\percent}). Bold marks the largest
and smallest gap and decoy capture; the last column is the share of wrong neutral answers on-decoy.}
\label{tab:mechanism}
\end{table}
\FloatBarrier

The three views agree on a mechanism-level split. Text, Duplicate, Prior, and Wrong Unit combine
frequent decoy hits with large reasoning-alignment gaps, consistent with selecting the wrong counting
process. Estimate has both the lowest decoy concentration and the smallest gap, indicating that its
remaining errors more often reflect perception or exact-enumeration capability. Structure and
Lookalike lie between these regimes, while the model-level variation cautions against treating the
ordering as a universal difficulty scale.

\clearpage
\subsection{Wording-Sensitivity Breakdown}
\begin{table}[h]
\centering
\small
\begin{tabular}{@{}lrrrrrr@{}}
\toprule
Subset & $n$ & Neutral & Visible & $\Delta$ & Repaired & Broken \\
\midrule
\textbf{All items} & \textbf{---} & \textbf{32.89} & \textbf{39.58}
                    & \textbf{+6.69} & \textbf{246} & \textbf{135} \\
\midrule
Duplicate  & 252 & 49.21 & 56.75 & +7.54  & 35 & 16 \\
Estimate   & 258 & 26.74 & 20.54 & -6.20  & 18 & 34 \\
Lookalike  & 291 & 33.68 & 38.49 & +4.81  & 35 & 21 \\
Prior      & 254 & 27.17 & 40.16 & +12.99 & 47 & 14 \\
Structure  & 257 & 32.30 & 31.52 & -0.78  & 31 & 33 \\
Text       & 222 & 27.03 & 46.85 & +19.82 & 51 & 7  \\
Wrong Unit & 228 & 35.09 & 43.86 & +8.77  & 41 & 21 \\
\bottomrule
\end{tabular}
\caption{Global wording sensitivity of Gemini-3.8-flash on the frozen release (\si{\percent}).
Family memberships overlap.}
\label{tab:global-wording-ablation}
\end{table}
\FloatBarrier

Wording is itself a selective alignment intervention, not a uniform clarification. Replacing
``are there'' with ``are visible'' strongly helps Text and Prior but harms Estimate, so no single
surface form resolves all mechanisms. This sensitivity motivates the multi-hop control: if richer
context were the missing ingredient, contextualized questions should repair failures more broadly.

\subsection{Cue Controls by Decoy Family}
\begin{table}[h]
\centering
\small
\setlength{\tabcolsep}{4pt}
\begin{tabular}{@{}lr cccc c cccc@{}}
\toprule
& & \multicolumn{4}{c}{\textbf{Gemini-3.8-flash}} & & \multicolumn{4}{c}{\textbf{GPT-6-astra}} \\
\cmidrule(lr){3-6}\cmidrule(lr){8-11}
\textbf{Decoy family} & \textbf{$n$} & \textbf{Neut.} & \textbf{Gen.} & \textbf{Mism.} & \textbf{Match.}
& & \textbf{Neut.} & \textbf{Gen.} & \textbf{Mism.} & \textbf{Match.} \\
\midrule
Duplicate  & 236 & 50.42 & 48.45 & 46.33 & \textbf{68.36} & & 26.41 & 24.58 & 28.53 & \textbf{71.61} \\
Wrong Unit & 221 & 33.94 & 34.69 & 35.29 & \textbf{42.08} & & 31.37 & 33.18 & 34.24 & \textbf{63.05} \\
Prior      & 243 & 28.26 & 36.90 & 33.20 & \textbf{49.25} & & 37.45 & 48.15 & 37.86 & \textbf{64.75} \\
Lookalike  & 272 & 31.50 & 30.88 & 33.70 & \textbf{38.97} & & 34.44 & 34.80 & 35.05 & \textbf{54.41} \\
Text       & 214 & 29.75 & 41.28 & 41.43 & \textbf{53.12} & & 52.49 & 64.17 & 62.93 & \textbf{75.70} \\
\midrule
Structure  & 246 & 29.54 & 28.05 & 30.49 & \textbf{34.82} & & 61.65 & 62.06 & 59.76 & \textbf{63.55} \\
Estimate   & 249 & 24.36 & 22.49 & 24.63 & \textbf{26.10} & & 59.30 & 58.77 & 57.83 & \textbf{61.31} \\
\bottomrule
\end{tabular}
\caption{Per-family accuracy under four cue conditions, two models and three repeats on the frozen
release (\si{\percent}). Bold marks the matched cue, the best condition in every row.}
\label{tab:cuefamily}
\end{table}
\FloatBarrier

Matched cues outperform both generic effort instructions and mismatched cues in every family, showing
that added deliberation alone is insufficient. Text is the partial exception in degree: even generic
or mismatched counting guidance suppresses some transcription shortcuts. Estimate barely responds to
any cue, reinforcing that the reasoning-alignment gap reflects process selection rather than residual capability.

\subsection{Tool Results by Decoy Family}
These are the raw and per-family results behind \Cref{sec:whentools,tab:tooluse}.
These tool studies retain their original evaluation cohorts; they have not been rescored on the
1,586-item release. Family sample sizes below refer to those original cohorts.
\begin{table}[h]
\centering
\small
\begin{tabular}{@{}llrrrr@{}}
\toprule
\textbf{Model} & \textbf{Arm} & \textbf{Neutral} & \textbf{Cued} & \textbf{$\rgap$} & \textbf{Rollouts} \\
\midrule
Qwen3.8-Flash-Next & bare       & 28.31 & 47.77 & 19.46 & 1 \\
Qwen3.8-Flash-Next & scaffold$^*$ & 21.67 & 41.39 & 19.72 & 1 \\
Qwen3.8-Flash-Next & $+$CI      & 30.00 & 56.63 & 26.63 & 1 \\
Qwen3.8-Flash-Next & \textbf{$+$CI$+$SAM3} & \textbf{33.61} & 55.72 & \textbf{22.11} & 1 \\
GPT-6-sol          & no tools   & 27.31 & 46.65 & 19.34 & 3 \\
GPT-6-sol          & $+$CI      & 35.72 & 57.65 & 21.93 & 1 \\
Claude-Opus-5.5    & no tools   & 32.91 & 44.88 & 11.97 & 3 \\
Claude-Opus-5.5    & $+$CI      & 31.75 & 46.20 & 14.45 & 1 \\
GPT-5.6-sol        & bare       & 18.98 & --- & --- & 1 \\
GPT-5.6-sol        & $+$Codex   & 20.72 & --- & --- & 1 \\
\bottomrule
\end{tabular}
\caption{All tool and no-tool experiments, with raw exact-match accuracy and reasoning-alignment
gap (accuracy in percent; gap in percentage points). CI denotes provider-native code execution (GPT \texttt{code\_interpreter};
Claude \texttt{code\_execution}); bold marks the fixed-agent CI+SAM3 arm. $^*$The scaffold exposes
no usable image tool; the earlier GPT-5.6 study reported only the full-set neutral arm.}
\label{tab:toolraw}
\end{table}

\begin{table}[h]
\centering
\small
\begin{tabular}{@{}lrrrr@{}}
\toprule
\textbf{Family} & \textbf{$n$} & \textbf{CI$\to$CI+SAM3} & \textbf{GPT-6 CI} & \textbf{Opus CI} \\
\midrule
Estimate    & 258 & \gapnarrows{-10.8} & \gapwidens{+1.2} & \gapwidens{+1.6} \\
Structure   & 257 & \gapnarrows{-8.6}  & \gapwidens{+4.1} & \gapwidens{+11.8} \\
Text        & 222 & \gapnarrows{-5.4}  & \gapwidens{+1.1} & \gapwidens{+0.6} \\
Prior       & 254 & \gapnarrows{-2.8}  & \gapwidens{+6.6} & \gapwidens{+1.3} \\
Lookalike   & 291 & \gapnarrows{-2.1}  & \gapwidens{+3.9} & \gapwidens{+3.5} \\
Wrong Unit  & 228 & \gapnarrows{-0.9}  & \gapnarrows{-0.8} & \gapwidens{+1.2} \\
Duplicate   & 252 & \gapnarrows{-2.8}  & \gapwidens{+1.1} & \gapnarrows{-1.7} \\
\midrule
\textbf{All} & --- & \gapnarrows{-4.52} & \gapwidens{+2.59} & \gapwidens{+2.48} \\
\bottomrule
\end{tabular}
\caption{Tool-induced change in the reasoning-alignment gap by family ($\rgapp$, percentage points).
CI$\to$CI+SAM3 holds the Qwen agent fixed; GPT-6 and Opus compare no tools with provider-native CI.
\textcolor{gapgreen}{\textbf{Green}} narrows the gap; \textcolor{gapred}{\textbf{red}} widens it.
Signed values preserve the distinction
in grayscale. No-tools values are three-rollout means; tool arms are one rollout.}
\label{tab:toolfamily}
\end{table}

The fixed-agent CI-to-CI+SAM3 comparison is the cleanest tool contrast: direct visual measurement
narrows the gap in every family, most strongly for Estimate and Structure. Provider-native CI, by
contrast, often increases both task performance and the gap because it also induces longer,
multi-turn deliberation; tool availability and agentic effort are therefore confounded. The tool arms
have one rollout, so these family effects are mechanism-level evidence rather than precise rankings.

\paragraph{Secondary agent surface.}
On the 301 items where GPT-5.6-sol's coding agent computes under both prompts, tool use gains
15.61 percentage points neutral and 12.62 points cued, so $\rgapp=-2.99$ points.
Selection affects the level, but the paired contrast agrees with CI$\to$CI+SAM3: gap reduction
depends on the tool surface rather than following automatically from higher tool-assisted accuracy.
\label{sec:secondary-agent}

\begin{figure}[h]
\centering
\includegraphics[width=.72\linewidth]{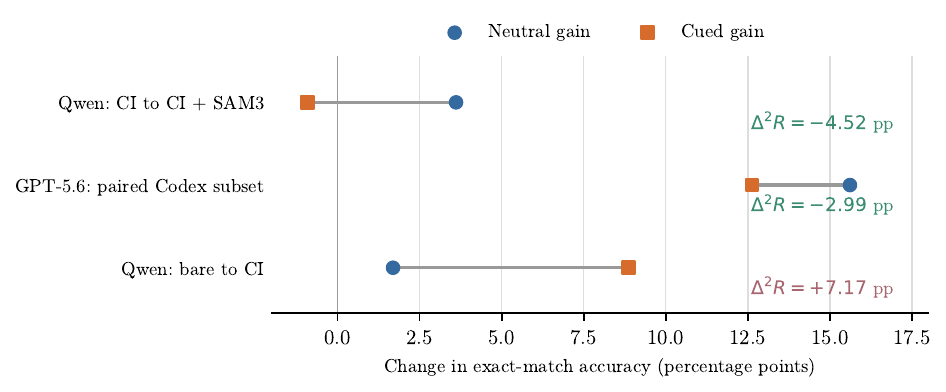}
\caption{\textbf{Tool effects depend on the agentic surface.} Blue circles and orange squares show
the tool-induced gains under neutral and cued prompts, respectively; their difference is
$\rgapp$, the change in the reasoning-alignment gap. Adding SAM3 to the same CI agent narrows the
gap by 4.52 points, and the coding-agent contrast narrows it by 2.99 points. The interpreter
contrast widens it by 7.17 points but also changes the agent, so it is not a tool-only comparison.
The Qwen contrasts retain their original evaluation cohort; Codex uses the 301-item
paired-computation subset. Neither has been rescored on the 1,586-item release.}
\label{fig:d2r}
\end{figure}
\FloatBarrier

\section{Run-to-run Variance}
\label{sec:variance}
The noise floor is measured \emph{before} any configuration is compared. Three independent repeats of
every model and condition on the frozen release give the standard deviations in \Cref{tab:noise}, and
treating two single runs as distinguishable requires a gap of roughly $1.96\sigma\sqrt{2}$. Every
comparison in this paper is reported against that threshold, and differences below it are described as
flat.

Pooling the within-cell variance over all 18 model-condition cells (36 degrees of freedom) gives
$\sigma = \SI{0.70}{\percent}$ and a minimum resolvable gap of \SI{1.93}{\percent}, which is what the
rest of the paper quotes. The pooled figure is preferred to any single cell: an estimate from three
repeats of one configuration carries two degrees of freedom and is too unstable to judge the rest of
the paper against.

Two features of the table matter for how it is used. The floor is \emph{condition-dependent} --- the cued arm is the quietest at \SI{1.57}{\percent}, against \SI{2.22}{\percent} neutral and
\SI{2.10}{\percent} multi-hop --- so a neutral-arm comparison needs a larger gap than a cued one to
clear the same bar. And it is \emph{model-dependent}: per-cell spread ranges from
\SI{0.19}{\percent} (GPT-6-astra multi-hop) to \SI{2.47}{\percent} (Kimi-k3 neutral), a factor of
thirteen. Kimi-k3 is the noisiest model in every condition and should never be read on a single run;
GPT-6-astra is the quietest, which is why its 22.11-percentage-point reasoning-alignment gap is the most secure
number in \Cref{tab:models}.

The repeats are independent draws at each model's own reference decoding rather than a seed sweep, so
this floor bounds sampling variation and not implementation variation. A seed-controlled arm would
bound the latter and has not been run.

\begin{table}[t]
\centering
\small
\begin{tabular}{@{}llccccc@{}}
\toprule
\textbf{Model} & \textbf{Condition} & \textbf{Mean} & \textbf{$\sigma$} & \textbf{Spread} & \textbf{Min.\ gap} & \textbf{Runs} \\
\midrule
GPT-6-astra            & Neutral   & 43.11 & 0.61 & 1.20 & 1.70 & 3 \\
                       & Cued      & 65.22 & 0.38 & 0.76 & 1.05 & 3 \\
                       & multi-hop & 42.70 & 0.04 & 0.07 & 0.11 & 3 \\
Gemini-3.8-flash       & Neutral   & 32.32 & 0.28 & 0.57 & 0.79 & 3 \\
                       & Cued      & 44.41 & 0.63 & 1.20 & 1.75 & 3 \\
                       & multi-hop & 32.43 & 0.79 & 1.43 & 2.19 & 3 \\
Qwen3.8-max            & Neutral   & 28.21 & 0.67 & 1.32 & 1.85 & 3 \\
                       & Cued      & 49.94 & 0.73 & 1.39 & 2.02 & 3 \\
                       & multi-hop & 32.79 & 0.72 & 1.43 & 1.99 & 3 \\
Gemini-3.1-pro-preview & Neutral   & 25.18 & 0.92 & 1.77 & 2.55 & 3 \\
                       & Cued      & 38.71 & 0.44 & 0.76 & 1.21 & 3 \\
                       & multi-hop & 27.23 & 0.28 & 0.54 & 0.79 & 3 \\
Claude-Fable-5.1       & Neutral   & 25.05 & 0.67 & 1.20 & 1.87 & 3 \\
                       & Cued      & 37.98 & 0.10 & 0.19 & 0.27 & 3 \\
                       & multi-hop & 27.69 & 0.72 & 1.43 & 1.99 & 3 \\
Kimi-k3                & Neutral   & 17.13 & 1.26 & 2.52 & 3.50 & 3 \\
                       & Cued      & 32.91 & 0.72 & 1.32 & 1.99 & 3 \\
                       & multi-hop & 20.88 & 1.20 & 2.24 & 3.31 & 3 \\
\midrule
\textbf{Pooled}        &           & ---   & \textbf{0.70} & --- & \textbf{1.93} & 36 d.o.f. \\
\bottomrule
\end{tabular}
\caption{Run-to-run variance, three independent repeats per cell
(\si{\percent}). Min.\ gap is $1.96\sigma\sqrt{2}$; the pooled \SI{1.93}{\percent} threshold is used
throughout.}
\label{tab:noise}
\end{table}

\section{Comparison to Other Benchmarks}
\label{app:comparison_to_countqa_fsc}
\begin{figure}[t]
\centering
\includegraphics[width=\linewidth]{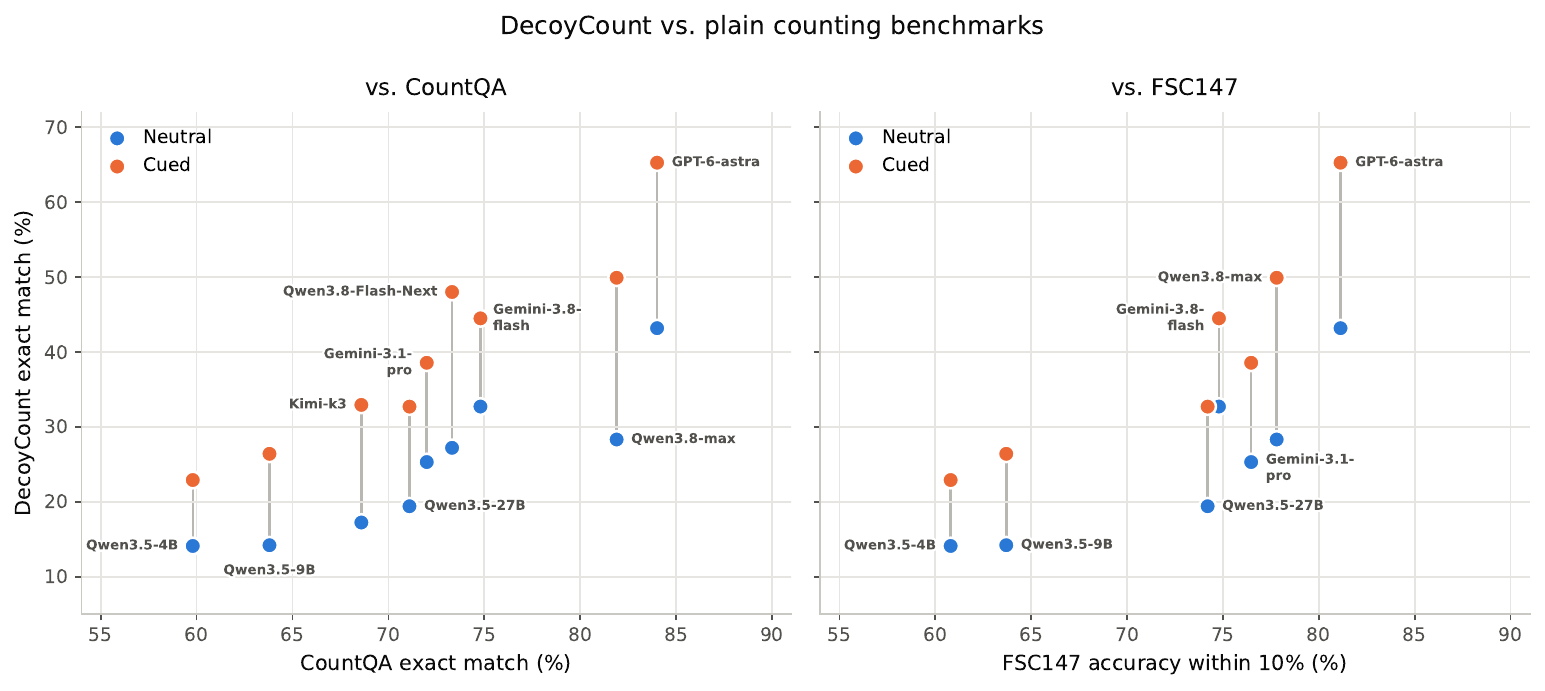}
\caption{\textbf{DecoyCount preserves broad counting rankings while exposing substantial new
headroom.} Each model is positioned by its CountQA exact match (left) or FSC147 within-10\%
accuracy (right), with DecoyCount exact match on the vertical axis. Blue and orange points are the
neutral and cued prompts; each connector therefore visualizes that model's reasoning-alignment gap.
Strong standard-counting scores coexist with low neutral DecoyCount accuracy, while the large cue
lifts show that part of the deficit is alignment rather than counting capability alone.
Points retain their original evaluation cohorts and have not been recomputed on the 1,586-item release.}
\label{fig:vs-counting}
\end{figure}

\Cref{fig:vs-counting} places DecoyCount against two plain counting benchmarks,
CountQA~\citep{tamarapalli2025countqa} (exact match) and FSC147~\citep{ranjan2021counting} (accuracy within
\SI{10}{\percent} of the true count), for every model with a number on both axes. The API models are
evaluated under the configuration of \Cref{sec:setup}; the Qwen3.5 counting numbers are single runs at
the models' reference decoding, and their DecoyCount numbers come from the scale study
(\Cref{sec:qwen35-scale-config}). Two things follow from the figure.

\paragraph{Benchmark headroom.} The best model on CountQA scores
\SI{84.0}{\percent} and on FSC147 \SI{81.1}{\percent}; the same model scores \SI{43.2}{\percent} on
DecoyCount neutral and \SI{65.3}{\percent} cued. Every model gives up between 41 and 54 points
between its CountQA score and its neutral DecoyCount score, and none recovers its counting number
even when the cue names the trap. The spread across models also widens: a 24-point range on CountQA
becomes 29 points on the neutral arm and 42 on the cued arm (slopes of 1.1 and 1.6 in
\Cref{fig:vs-counting}), so DecoyCount separates models that the counting benchmarks pack into a
few points.

\paragraph{External validity.} Better counting still
buys DecoyCount accuracy. Across the nine models with CountQA scores the rank correlation with
DecoyCount is $\rho = 0.98$ on the neutral arm and $0.97$ cued; against FSC147 it is $0.83$ and
$0.94$. The relation is monotone rather than flat or inverted: the Qwen3.5 sizes climb both axes in
order, and GPT-6-astra leads all four. What the decoys add is a second axis on top of counting
skill.


\section{Additional multi-hop results}

The condition remains count-bottlenecked: downstream accuracy is 24.43--48.10 percentage points
higher when the neutral count is correct (\Cref{tab:mhprop}). Rescue concentrates in unit-settling
families (\SI{24.0}{\percent} versus \SI{13.7}{\percent}; \Cref{tab:mhfamily}), so context can settle
units but cannot replace alignment guidance. The main contextualization result is in \Cref{sec:mh}.

\begin{table}[t]
\centering
\small
\begin{tabular}{@{}lccccc@{}}
\toprule
& \multicolumn{2}{c}{\textbf{Neutral count correct}} & \multicolumn{2}{c}{\textbf{Neutral count wrong}} & \\
\cmidrule(lr){2-3}\cmidrule(lr){4-5}
\textbf{Model} & \textbf{multi-hop acc.} & \textbf{$n$} & \textbf{multi-hop acc.} & \textbf{$n$} & \textbf{Gap} \\
\midrule
GPT-6-astra       & 70.30 & \phantom{0}646 & 21.05 & \phantom{0}824 & $+$49.26 \\
Gemini-3.8-flash  & 62.34 & \phantom{0}489 & 17.50 & \phantom{0}981 & $+$44.84 \\
Qwen3.8-max       & 66.00 & \phantom{0}426 & 19.24 & 1044 & $+$46.76 \\
Gemini-3.1-pro    & 45.12 & \phantom{0}380 & 20.98 & 1090 & $+$24.14 \\
Claude-Fable-5.1  & 60.73 & \phantom{0}378 & 16.25 & 1092 & $+$44.49 \\
Kimi-k3           & 54.24 & \phantom{0}256 & 13.85 & 1214 & $+$40.39 \\
\bottomrule
\end{tabular}
\caption{Accuracy under the multi-hop condition, conditioned on the paired neutral count, three repeats
(\si{\percent}); $n=1470$ (see Exclusions, \Cref{sec:impl}).}
\label{tab:mhprop}
\end{table}

\begin{table}[t]
\centering
\small
\setlength{\tabcolsep}{4pt}
\begin{tabular}{@{}lcccccc@{}}
\toprule
\textbf{Decoy family} & \textbf{GPT-6} & \textbf{G-3.8-fl.} & \textbf{Qwen-max} & \textbf{G-3.1-pro} & \textbf{Fable-5.1} & \textbf{Kimi-k3} \\
\midrule
Duplicate   & $+$12.2 & \phantom{0}$+$5.3 & $+$16.6 & $+$24.9 & \phantom{0}$+$8.6 & \phantom{0}$+$9.2 \\
Wrong Unit  & \phantom{0}$+$1.8 & \phantom{0}$-$2.9 & \phantom{0}$+$6.8 & \phantom{0}$+$8.9 & \phantom{0}$+$1.6 & \phantom{0}$+$5.3 \\
Lookalike   & \phantom{0}$+$0.1 & \phantom{0}$+$2.1 & \phantom{0}$+$4.2 & \phantom{0}$+$5.3 & \phantom{0}$+$4.4 & \phantom{0}$+$1.6 \\
Text        & $-$15.7 & \phantom{0}$-$5.8 & \phantom{0}$-$0.5 & $-$15.2 & \phantom{0}$-$9.0 & \phantom{0}$+$1.8 \\
Prior       & \phantom{0}$+$0.1 & \phantom{0}$+$0.1 & \phantom{0}$+$3.9 & \phantom{0}$-$7.9 & \phantom{0}$+$2.9 & \phantom{0}$+$3.3 \\
Structure   & \phantom{0}$-$8.3 & \phantom{0}$-$3.0 & \phantom{0}$-$3.3 & \phantom{0}$-$4.6 & \phantom{0}$+$2.8 & \phantom{0}$+$1.9 \\
Estimate    & \phantom{0}$-$3.2 & \phantom{0}$-$4.0 & \phantom{0}$-$2.5 & \phantom{0}$-$4.4 & \phantom{0}$+$0.0 & \phantom{0}$+$1.0 \\
\midrule
\textbf{Overall} & \phantom{0}$-$1.27 & \phantom{0}$-$0.86 & \phantom{0}$+$3.83 & \phantom{0}$+$1.36 & \phantom{0}$+$1.95 & \phantom{0}$+$3.45 \\
\bottomrule
\end{tabular}
\caption{Difference between multi-hop and neutral accuracy by family and model, three repeats (\si{\percent});
$n=1470$ (see Exclusions, \Cref{sec:impl}). The first three rows are unit-settling families.}
\label{tab:mhfamily}
\end{table}

\begin{table}[t]
\centering
\small
\begin{tabular}{@{}lrrrrr@{}}
\toprule
& \multicolumn{2}{c}{\textbf{Rates}} & \multicolumn{3}{c}{\textbf{Contribution to the delta}} \\
\cmidrule(lr){2-3}\cmidrule(lr){4-6}
\textbf{Decoy family} & \textbf{Lost} & \textbf{Rescued} & \textbf{Cost} & \textbf{Gain} & \textbf{Net} \\
\midrule
Duplicate  & 34.4 & \textbf{32.7} & $-$10.7 & $+$22.6 & $+$11.9 \\
Wrong Unit & 42.0 & 17.9 & $-$10.2 & $+$13.5 & \phantom{0}$+$3.3 \\
Lookalike  & 44.5 & 20.9 & $-$12.5 & $+$15.1 & \phantom{0}$+$2.6 \\
Text       & \textbf{52.1} & \textbf{11.9} & $-$16.3 & \phantom{0}$+$8.2 & \phantom{0}$-$8.1 \\
Prior      & 43.5 & 14.0 & $-$10.3 & $+$10.7 & \phantom{0}$+$0.4 \\
Structure  & 37.7 & 14.8 & $-$12.9 & \phantom{0}$+$9.7 & \phantom{0}$-$3.2 \\
Estimate   & 37.4 & 13.9 & $-$12.5 & \phantom{0}$+$9.2 & \phantom{0}$-$3.3 \\
\bottomrule
\end{tabular}
\caption{Transitions under the multi-hop condition by family, averaged over six models (\si{\percent}). With neutral
accuracy $A$, survival $R$, and rescue $W$, $\Delta=-A(1-R)+(1-A)W$.}
\label{tab:mhdecomp}
\end{table}

The decomposition explains why the aggregate multi-hop change is small despite large item-level
movement. Context rescues Duplicate cases, where an explicit real-world unit can settle whether a
depiction is a new entity, but it loses many Text cases and provides little net help for Structure or
Estimate. Moreover, downstream accuracy remains 24--48 points higher when the initial count is
correct. Richer context therefore cannot generally compensate for a misaligned or incorrect visual
count; its gains and losses mostly cancel, whereas mechanism-matched cues directly target process
selection.